\documentclass[conference]{ieeeconf}      % Use this line for a4 paper
\pdfoutput=1

\IEEEoverridecommandlockouts                              % This command is only needed if 
\usepackage{graphicx}
\usepackage{subfig}
\usepackage{amsmath} % assumes amsmath package installed
\usepackage{amssymb}  % assumes amsmath package installed
\usepackage{algorithmic, algorithm}
\usepackage{xcolor}

\usepackage{pifont}% http://ctan.org/pkg/pifont
\newcommand{\xmark}{\ding{55}}%

\title{\LARGE \bf
Learning to Solve a Rubik's Cube with a Dexterous Hand
}

\author{Tingguang Li$^\dagger$, Weitao Xi, Meng Fang, Jia Xu, Max Q.-H. Meng*, \emph{Fellow}, \emph{IEEE} 
\thanks{$^\dagger$Work mainly done during an internship at Tencent AI Lab.}
\thanks{*Corresponding author.}
\thanks{Tingguang Li and Max Q.-H. Meng are with the \textit{Robotics, Perception and AI Laboratory} at Electronic Engineering Department, The Chinese University of Hong Kong, Hong Kong SAR, China. Weitao Xi, Meng Fang and Jia Xu are with Tencent AI Lab, Shenzhen, China. {\tt\small email: \{tgli, qhmeng\}@ee.cuhk.edu.hk},  \tt\small \{weitaoxi, mfang, jiajxu\}@tencent.com.}
}
\begin{document}

\makeatletter
\g@addto@macro\@maketitle{
  \begin{figure}[H]
  \setlength{\linewidth}{\textwidth}
  \setlength{\hsize}{\textwidth}
 % \vspace{-2mm}
  \centering
 	\begin{tabular}{@{}c@{\hspace{1mm}}c@{\hspace{1mm}}c@{\hspace{1mm}}c@{}}
  \includegraphics[width=0.245\textwidth]{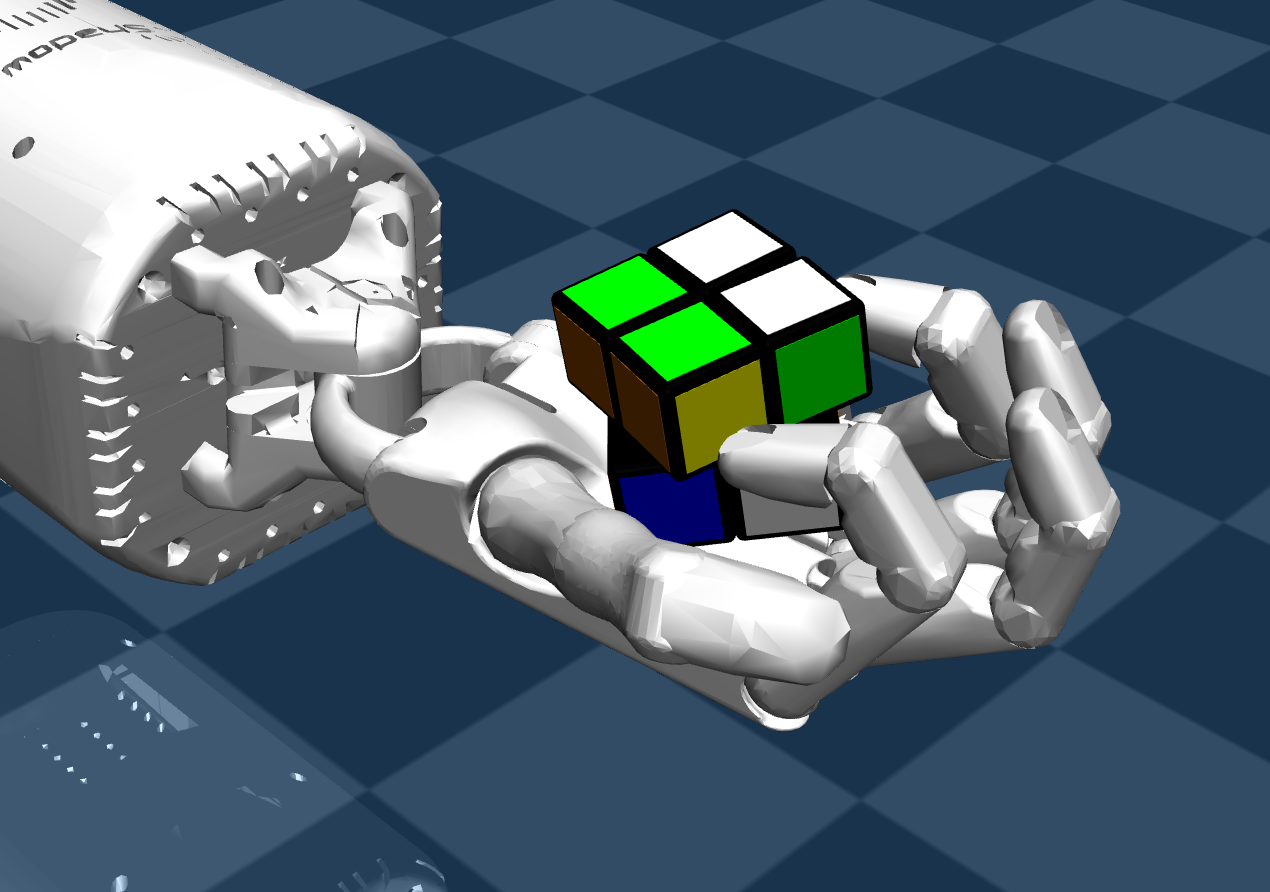} &  
  \includegraphics[width=0.245\textwidth]{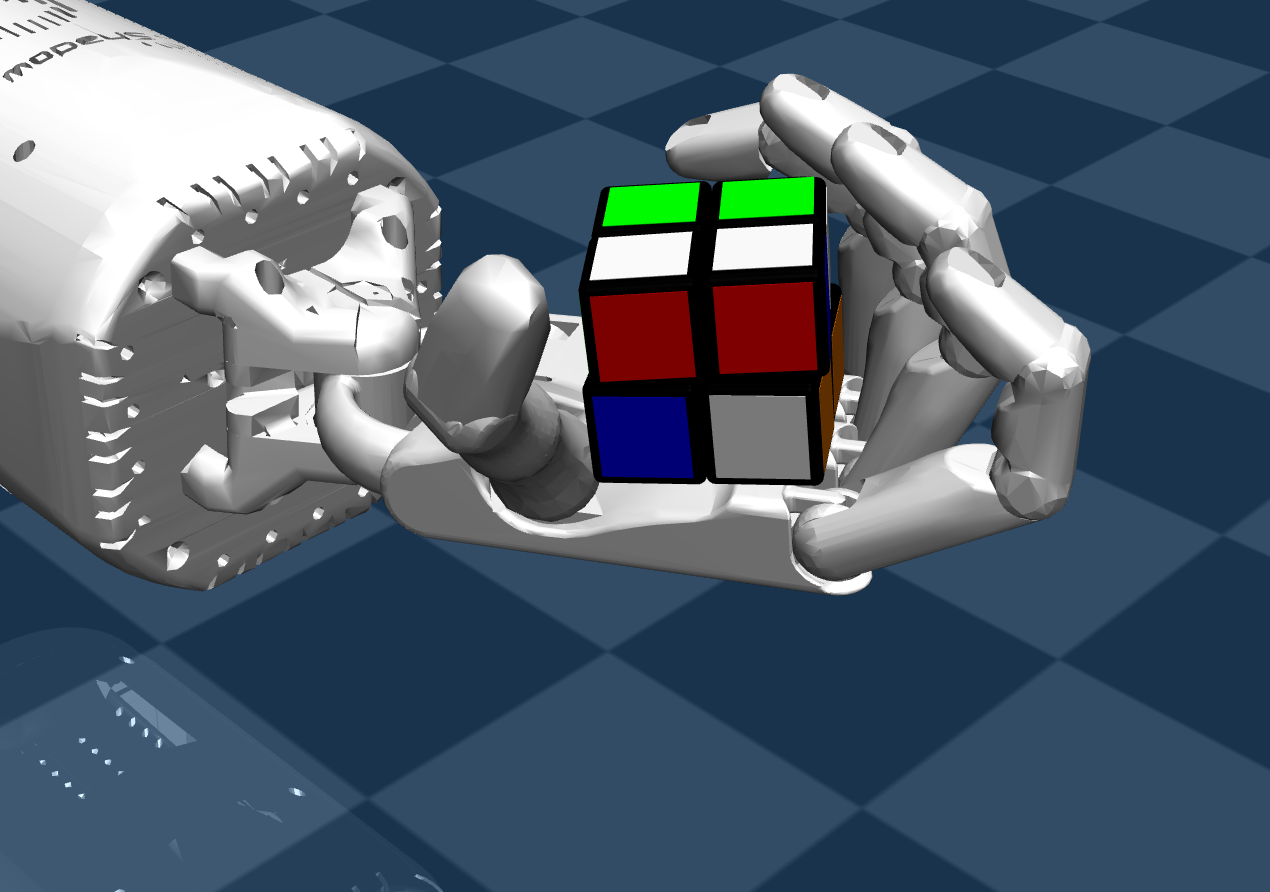} &  \includegraphics[width=0.245\textwidth]{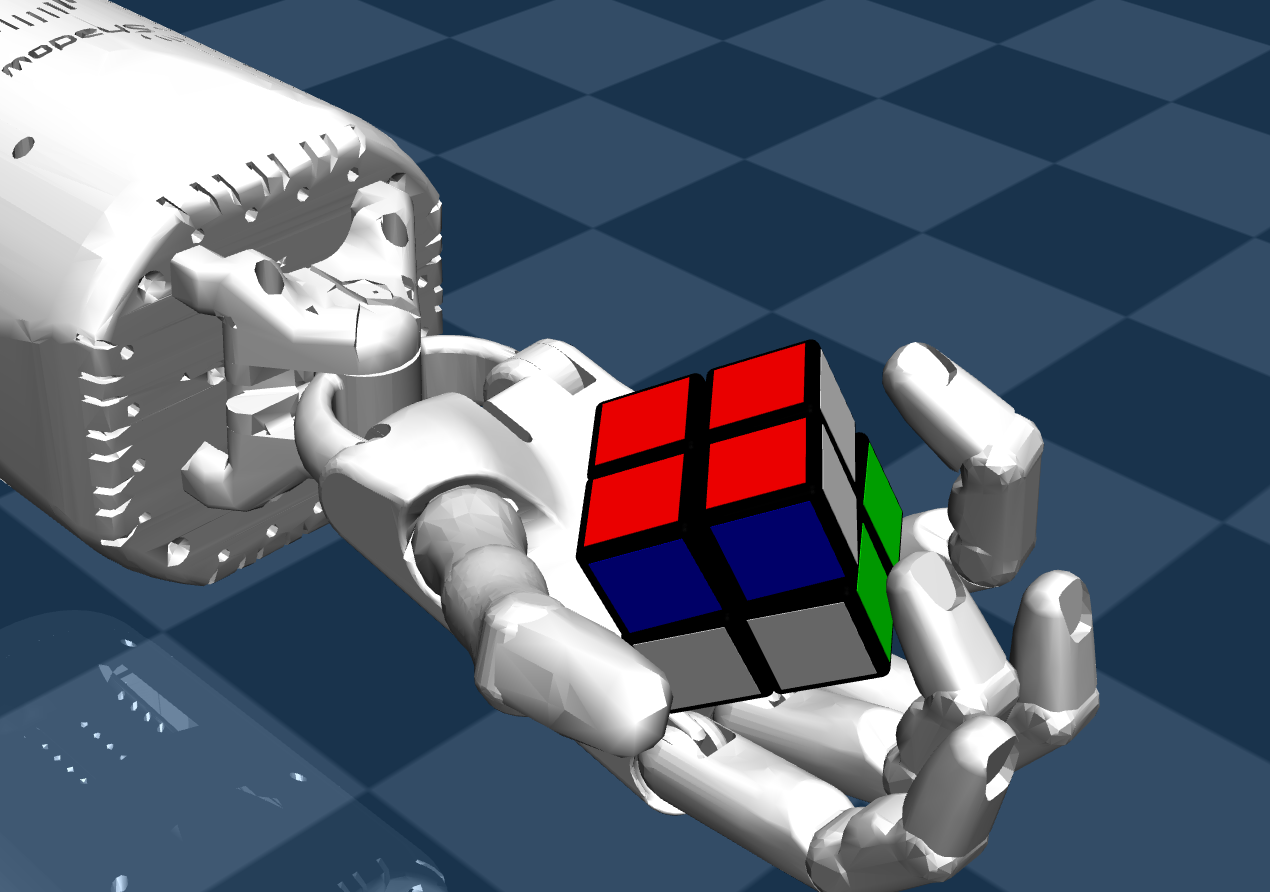} &
  \includegraphics[width=0.245\textwidth]{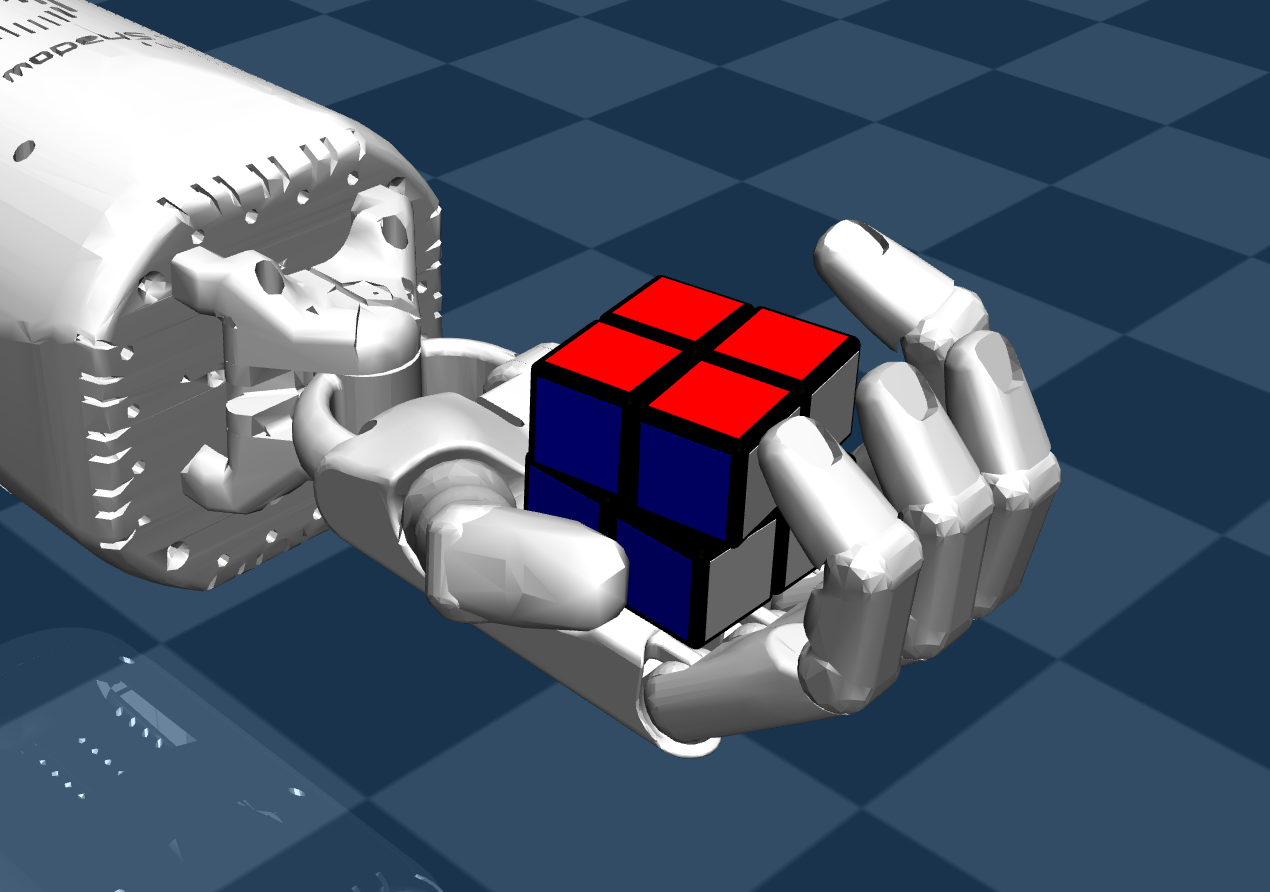} \\
  (a) &(b) & (c) &(d)
 	\end{tabular}
  %\vspace{-2mm}
  \caption{Our five-fingered dexterous hand solves a scrambled Rubik's Cube by operating its layers and changing its pose. Our method starts with a random state (a), plans the optimal move sequence (b,c), and reaches the desired state (d).}
  \label{fig:task}
  %\vspace{2mm}
  \end{figure}
}
\makeatother

\maketitle
\thispagestyle{empty}
\pagestyle{empty}

%%%%%%%%%%%%%%%%%%%%%%%%%%%%%%%%%%%%%%%%%%%%%%%%%%%%%%%%%%%%%%%%%%%%%%%%%%%%%%%%
\begin{abstract}
% The dexterity of multi-fingered hands has been investigated to accomplish a series of tasks like rotating cubes, using tools and relocating objects. 
% However, complex tasks which involve multiple steps and diverse internal structure of objects have less been addressed. In this paper, we focus on solving a 2x2x2 Rubik's Cube, an object containing high dimensional state space, with a 24-DoF robot hand. We propose a hierarchical structure combining model-based and model-free approaches to separate planning and manipulation. A model-based cube solver finds an optimal move sequence for restoring the cube and a model-free cube operator controls the hand to execute each move step by step. The cube operator is further decomposed into two stages, separately trained by DRL. We demonstrate the effectiveness of our method by solving $1400$ randomly scrambled Rubik's cubes using a Shadow Hand in simulated experiments and it achieves an average success rate of $90.3\%$. 
We present a learning-based approach to solving a  Rubik's cube with a multi-fingered dexterous  hand. Despite the promising performance of dexterous in-hand manipulation, solving complex tasks which involve multiple steps and diverse internal object structure has remained an important, yet challenging task. In this paper, we tackle this challenge with a hierarchical deep reinforcement learning method, which  separates planning and manipulation. A model-based cube solver finds an optimal move sequence for restoring the cube and a model-free cube operator controls all five fingers to execute each move step by step.
To train our models, we build a high-fidelity simulator which manipulates a Rubik's Cube, an object containing high-dimensional state space, with a $24$-DoF robot hand. Extensive experiments on $1400$ randomly scrambled Rubik's cubes demonstrate the effectiveness of our method, achieving an average success rate of $90.3\%$.

\end{abstract}

%%%%%%%%%%%%%%%%%%%%%%%%%%%%%%%%%%%%%%%%%%%%%%%%%%%%%%%%%%%%%%%%%%%%%%%%%%%%%%%%
\section{INTRODUCTION}
% moe: the background of dexterous hands and its difficulty

Dexterous in-hand manipulation is a key building block for robots to achieve  human-level dexterity, and accomplish everyday tasks which involve rich contact. Despite concerted progress, reliable multi-fingered  dexterous hand manipulation has remained an open challenge, due to its  complex contact patterns, high dimensional action space, and fragile mechanical structure. 

%Due to versatility and anthropomorphic nature, multi-fingered dexterous hands are believed to have the potential to achieve human-level dexterity, and accomplish everyday tasks which involve rich contact.
%On the other hand, its high dimensional action space, complex contact patterns and fragile mechanical structure make it a challenging problem.

% Current investigations on dexterous manipulation with multi-fingered hands mainly treat the target object as a whole and focus on its external goal, like object re-location and re-pose \cite{andrychowicz2018learning} \cite{rajeswaran2017learning}, or the relationship between objects, like tool use \cite{rajeswaran2017learning}, or the relationship between objects and environments \cite{chavan2018hand}.
%while manipulating objects with diverse internal states is less addressed.

Traditional approaches \cite{mordatch2012contact,bai2014dexterous} studied simple tasks including grasping and picking up objects in simulation.
Recent deep reinforcement learning (DRL) based methods have demonstrated promising performance 
on dexterous manipulation of object re-location and re-posing \cite{andrychowicz2018learning,rajeswaran2017learning}, or  tool using and door opening \cite{rajeswaran2017learning,chavan2018hand}.
However, all these methods  only consider external states and short interactions (e.g., interactions between objects, or interactions between objects and environments), lacking key capabilities to solve complex tasks with diverse internal states. 

% for autonomous robots, it is still challenging for complex manipulation tasks, such as solving the Rubik's Cube, as shown in Fig.~\ref{fig:task}. Recent work, such as~\cite{andrychowicz2018learning}, has shown promising in-hand manipulation results in simulation and physical robots. It rotates an object, e.g. a box, into a specific location and rotation. Prior works \cite{mordatch2012contact} and \cite{bai2014dexterous} studied grasping and picking up objects in simulation. However, unlike these tasks, solving the Rubik's Cube needs to manipulate objects with diverse internal states. These works do not work well on tasks with this level of difficulty.

In this paper, we focus on solving the Rubik's Cube (shown in Fig.~\ref{fig:task}), an object with extremely varied states, using a Shadow Dexterous Hand. The Rubik's Cube is a classic 3-Dimensional combination puzzle with 6 faces and each face can be twisted $90^\circ$. The goal is to reach the state where all stickers on each face have the same color. Considering the mechanical capacity of the Shadow Hand,
%that it may be too challenging to manipulate a 3x3x3 or higher order Rubik's Cube, 
in this work we focus on solving the simplest case of the Rubik's Cube, the 2x2x2 Rubik's Cube, though our solution can be applied to higher order Rubik's Cubes.

\begin{figure*}
     \centering
     \includegraphics[width=\textwidth]{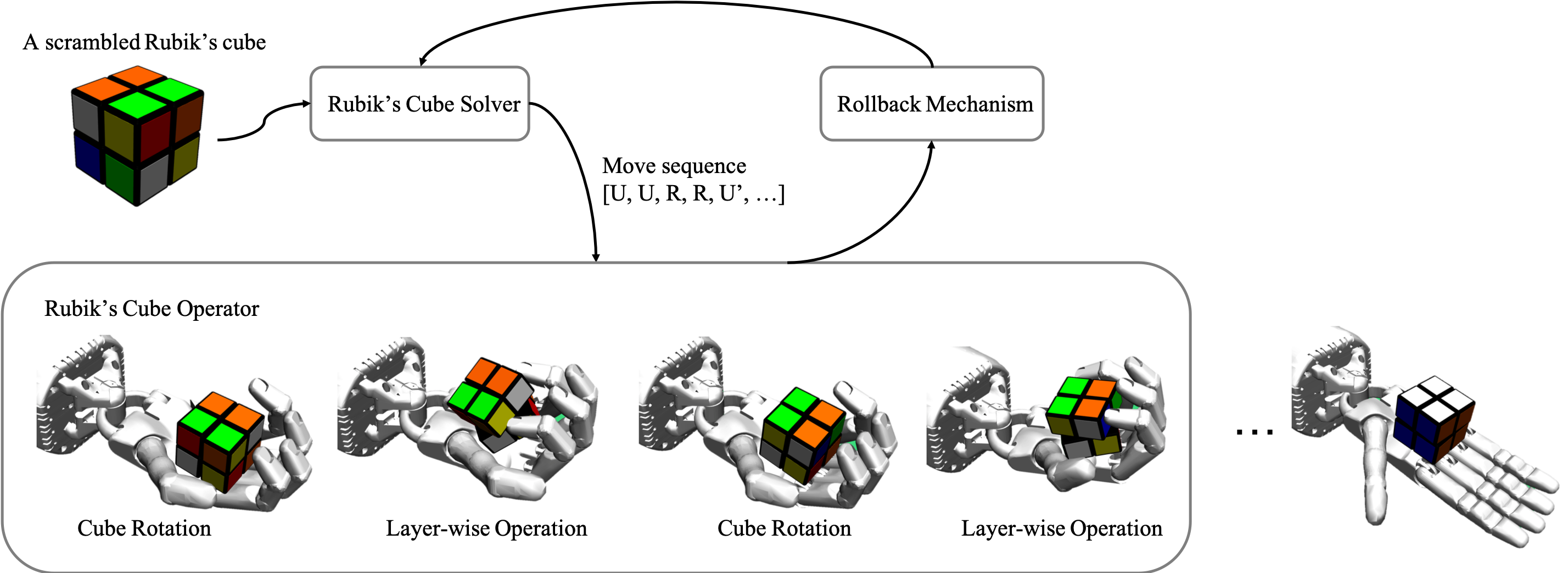}
     \caption{Overall structure. Given a randomly scrambled Rubik's Cube, the Rubik's Cube Solver finds a move sequence and Rubik's Cube Operator executes each move by rotating the cube and operating its layer sequentially. The rollback mechanism additionally checks the status of each move and provides feedback signals.}
     \label{fig:pipeline}
\end{figure*}

One distinguishing characteristic of this task is that it involves planning in high-dimensional state space and controlling in high-dimensional action space. A 2x2x2 Rubik's Cube has $3,674,160$ internal configurations \cite{frey1982handbook} in total and a Shadow Hand has 24-DoF. This makes it extremely challenging to directly employ DRL  algorithms. To this end, we decompose the whole task into a hierarchical structure combining model-based and model-free approaches to separate the planning and manipulation. 
First, a model-based Rubik's Cube Solver finds an optimal move sequence for restoring the cube. 
Second, a model-free Cube Operator controls the hand to execute each move step by step. The cube operator consists of two atomic actions, cube rotation and layer-wise operation, separately trained by DRL. In this way, each move is executed by re-posing the cube and operating its layers. To further improve the system stability, a rollback mechanism is developed to check the status of each move.
The overall structure is depicted in Fig. \ref{fig:pipeline}. 

%\iffalse
%\begin{figure}
 %    \centering
 %    \includegraphics[width=240pt]{./fig/rubik_cube.png}
 %    \caption{Rubik's Cube and Shadow Hand in Mujoco simulator.}
 %    \label{fig:task}
 %\end{figure}
 % \fi
 
%To overcome these two issues, we propose a joint training method where the two adjacent stages are connected and their goal are shared.

To thoroughly evaluate our method, we have developed a high-fidelity simulator with a Shadow hand to manipulate a  Rubik's Cube, using the Mujoco engine \cite{todorov2012mujoco}. We optimize the performance of our simulator to produce hundreds of frames per second on a single CPU core. This enables us to train a model within one day without GPUs.
We demonstrate the effectiveness of our method by solving $1400$ randomly scrambled Rubik's Cubes using the Shadow hand and our method achieves an average success rate of $90.3\%$.
Our simulator and source code will be publicly available to advance future research.
 
\section{RELATED WORK}
\subsection{Dexterous In-Hand Manipulation}
Dexterous in-hand manipulation is a new area in dexterous manipulation. Bai et al. \cite{bai2014dexterous} propose a palm controller and a multi-finger controller for rotating objects on hand to desired positions. Mordatch et al. \cite{mordatch2012contact} use trajectory optimization method to grab and move a pen-like object. Kumar et al. \cite{kumar2016optimal} learn a control policy and implement the policy on a physical humanoid robot hand.
There is also significant progress using deep reinforcement learning in dexterous in-hand manipulation. Barth et al. \cite{barth2018distributed} and Andrychowicz et at. \cite{andrychowicz2017hindsight} use deep reinforcement learning to rotate different objects, such as cubes, balls, pen-like objects in simulation environments. Andrychowicz et at.\cite{andrychowicz2018learning} also implement their methods on a physical Shadow Dexterous Hand.

\subsection{Hierarchical Reinforcement Learning}
Using hierarchies of policies to solve complex tasks is popular in reinforcement learning. In standard hierarchical reinforcement learning \cite{kulkarni2016hierarchical}, \cite{heess2016learning}, \cite{nachum2018data}, \cite{li2018learning} one higher-level policy and one lower-level policy policy are trained. The output of higher-level policy is the goal of lower-level policy. For example, in the task of controlling a robot to escape from a maze, the lower-level policy learns how to move the robot on a certain direction and the higher-level policy learns the path to escape from the maze.

\subsection{Rubik's Cube}
There are much research on solving Rubik's Cubes with a minimal number of moves. It has been proven that a 2x2x2 cube can be solved within 11 moves using half turn metric or 14 moves using quarter turn metric \cite{frey1982handbook}. Heuristic based search algorithms have been employed to find optimal solutions. Korf \cite{korf1997finding} use a variation of the A$^*$ heuristic search, along with a pattern database heuristic to find the shortest possible solutions. As for 3x3x3 Rubik's cubes, the minimal number of steps is still an open problem. But the minimal number is between 17 and 20 \cite{rokicki2014diameter}, \cite{kunkle2009harnessing}.

At the same time, there are attempts to solve a physical Rubik's Cube with robots. Zielinski et al. \cite{zielinski2007rubik} use two robot hands with a specially designed gripper to solve the Rubik's Cube. Rigo et al. \cite{rigo2018rubik} build a high-speed 3-fingered robot hand capable to solve a 3x3x3 Rubik's Cube. However, there are little work focusing on solving the Rubik's Cube with one general-purpose multi-fingered robot hand.

\section{METHODOLOGY} \label{sec:method}
 \subsection{Overview}
 Multi-fingered robot hands have a potential capability for achieving dexterous manipulation of objects. 
 Compared with special-purpose devices optimized for single specialized operation, multi-fingered manipulators are suitable for a wide range of tasks. However, it is still challenging to apply them to many applications due to their complexity of control \cite{siciliano2016springer}. One possible solution is employing deep reinforcement learning \cite{andrychowicz2018learning} where control policies are obtained in a trial-and-error manner, avoiding explicitly calculating its complex dynamics.

However, directly applying DRL to solve a Rubik's Cube using a dexterous Shadow Hand is challenging due to the following characteristics of the task: high-dimensional state space and high-dimensional action space.
 (1) For the high-dimensional state space, a 2x2x2 Rubik's Cube has $3,674,160$ states with only one goal state. This results in an inefficient exploration, i.e. the reward is so sparse that the agent may never solve the cube and thus never receive a learning signal.
 (2) For the high-dimensional action space, the Shadow Hand has 24-DoF with 20 actuated joints, making the learning process even more difficult.
 Therefore, a structure that can improve exploration efficiency while keeping the flexibility of model-free methods is necessary.
 
 %In this paper, 
 To address these challenges, we present a hierarchical structure that integrates model-based methods to plan a move sequence for solving the Rubik's Cube and model-free methods to control the hand to execute the planned moves. Given a randomly shuffled Rubik's Cube, the model-based part, Rubik's Cube Solver, finds a move sequence for solving the cube. Then the model-free part, Cube Operator, controls the robot hand to execute the corresponding moves sequentially. 
 We develop a 2-stage Cube Operator and define two atomic actions, cube rotation and layer-wise operation, respectively. At the first stage, the cube is rotated to a right pose and at the second stage, some specific layers are twisted by the hand.  
 To further increase the system stability, a rollback mechanism is proposed to make the system close-loop. 
 
 \subsection{Rubik's Cube Representation}
 The 2x2x2 Rubik's Cube is made of $8$ smaller cubes referred to as cubelets. Each cubelet has three exposed faces, called facelet, attached with stickers and there are $24$ facelets in total. Therefore, the cube state, denoted as $s$, can be represented as a $24$-dim vector describing the sticker color on each facelet. The 2x2x2 Rubik's Cube consists of $3,674,160$ states with only one goal state $s_{goal}$ where all the stickers on each side of the cube are of the same color. 
 %A frame, denoted as $\mathcal{F}_1$, is attached to the cube where its origin locates at the center of the cube and its orientation is fixed with one particular cubelet.
 
 Understanding the basic move notations is essential to understand the process of restoring the cube. Standard notations for the quarter turn metric, which is used in our work, are Up ($U$, $U'$), Down ($D$, $D'$), Right ($R$, $R'$), Left ($L$, $L'$), Front ($F$, $F'$) and Back ($B$, $B'$), each designating one of the cube layers rotating $90^\circ$ or $90^\circ$ clockwise. We denote applying a move, for example $U$, to a cube $C$ as $C*U$.  Accordingly, a scrambled Rubik's Cube $C$ can be restored by applying a sequence of moves, e.g., $C*U*F*R$. 
 
 \subsection{Rubik's Cube Solver} \label{sec:solver}
 Although there are attempts to calculate the move sequence for solving the Rubik's Cube without human knowledge \cite{mcaleer2018solving}, the success rate as well as the solving time are not guaranteed. Considering the fact that there are some well-developed algorithms for calculating the move sequence, we employ a model-based Rubik's Cube solver in this part.
 
 Different computational algorithms are proposed for solving the Rubik's Cube, such as Thistlewaite's algorithm \cite{thistlethwaite1981thistlethwaite}, Kociemba's algorithm \cite{kociemba} and Iterative Deepening A* (IDA*) search algorithm \cite{korf1997finding}. 
 In our architecture, we employ IDA* algorithm that can find an optimal sequence to restore a randomly scrambled Rubik's Cube. IDA* is a depth-first search algorithm, that looks for increasingly longer solutions in a series of iterations. It uses a lower-bound heuristic to prune its branches once a lower bound on their length exceeds the current iteration bound. Given a randomly scrambled cube $C$, we apply $1$ move and calculate all the result cubes $\{C_1\}$. If the goal state is not in $\{C_1\}$, the cube set $\{C_2\}$ is generated by additionally applying 1 move to cubes in $\{C_1\}$, i.e. $C*U*U$, $C*U*R$, $C*U*F$, etc. If some cubes are found involved too many moves that exceed the optimal bound, which is $14$ for the 2x2x2 Rubik's Cube, they are removed from the search path. This process repeats until the goal state is found. Then the move sequence with the shortest length to the goal state is recorded as the optimal path. There may exist multiple optimal paths for the same scrambled cube and only path one is selected for operating.

 \subsection{Rubik's Cube Operator} \label{sec:atomic_actions}
 Rubik's Cube operator receives the planned move sequence from the cube solver and controls the Shadow Hand to execute moves sequentially. First, we need to find out what is the best strategy to execute moves. Currently the quarter turn set contains 12 moves, i.e. $\{U,U',D,D',R,R',L,L',F,F',B,B'\}$, which requires 12 actions accordingly. However, it is not necessary to implement all the moves. For example, $U$ and $D'$ achieve the same result in terms of solving a Rubik's Cube. Therefore, the quarter turn set can be represented as $\{U,U',R,R', F,F'\}$.
 
 Another observation is the difficulty level of each move varies in terms of the Shadow Hand. Due to the kinematics and joint limits of the hand, some moves are naturally more difficult to achieve. For example, performing $F$ is harder than $U$ since it is inconvenient for the fingers to touch specific locations of the front layer. Apart from the layer types, the move directions also contribute to system reliability. For example, it is easier for the Shadow Hand to execute $U'$ than $U$ since the little finger is longer and more flexible compared with the thumb. To quantitatively compare the 6 moves, we separately train 6 models with the goal of $90^\circ$ or $-90^\circ$ between two specific layers and the result is discussed it in Section \ref{sec:exp_stage2}. It is found that the move $U'$ is much more reliable than all other moves with a higher success rate. 
 
 Aiming to make the system more stable, we only choose the move $U'$ and decompose the cube operator into two stages. At the first stage, the hand rotates the whole cube to a specific pose so that the desired move can be achieved by performing $U'$. For example, if the desired move is $R'$, we first rotate the cube so that the right layer is on the top. At the second stage, we just perform $U'$ to the up layer. Besides, we perform $U'$ for 3 times for the cases where $U$ is required. The detailed introduction is as follows.
 
 \subsubsection{Stage 1. Cube Rotation}
 The goal of this stage is to re-pose the whole cube so that the target layer is on the top. It is not necessary to enable our model to rotate all the faces to the top, due to the fact that rotating the same angle to the counter faces on the top has the same result. Therefore, three poses are sufficient for our task. Formally, the goal is composed of a target position $x'$ and a target orientation $q'$. The mapping from the desired move $m$ to the target orientation (in quaternions) is
  \[
 q' = \begin{cases}
 [1, 0, 0, 0] & \text{if $m\in\{U,U'\}$,} \\
 [0.707, 0, -0.707, 0] & \text{if $m\in\{R,R'\}$,} \\
 [0.707, 0.707, 0, 0] & \text{if $m\in\{F,F'\}$.} \\
 \end{cases}
 \]
 
 The goal is reached when $|x'-x|_2<\delta_x$ and $2*\text{arccos}(\text{Real}(q'*\text{conj}(q)))<\delta_q$, where $x$ is the achieved position, $q$ is the achieved orientation, $\delta_x$ and $\delta_q$ are position and orientation threshold.
 
 \subsubsection{Stage 2. Layer-wise Operation}
 After the target layer is rotated on the top by Stage 1, $U'$ is performed to rotate the top layer for $-90^\circ$. The goal is the angle between the up layer and down layer and it is achieved when $|\theta - \theta'| < \delta_\theta$, where $\theta$, $\theta'$ and $\delta_\theta$ are achieved angle, target angle and angle threshold.
 
 \subsection{Training Atomic Actions}
 We use Hindsight Experience Replay (HER) \cite{andrychowicz2017hindsight} to train atomic actions. We have two atomic actions for Stage 1 and Stage 2, respectively. Both atomic actions are very difficult and often fail. Thus we consider using HER to learn from failures in order to be more efficient.
 HER is a simple and effective reinforcement learning method of manipulating the replay buffer used in off-policy RL algorithms that allows it to learn policies more efficiently from sparse rewards. It assumes the goal being pursued does not influence the environment dynamics.
 After experiencing an episode $\{s_0,  s_1, \cdots, s_{T}\}$, every transition $s_t \rightarrow s_{t+1}$ along with the goal for this episode is usually stored in the replay buffer. Some of the saved episodes fail to reach the goal, providing no positive feedback to the agent.
 However, with HER, the failed experience is modified and also stored in the replay buffer in the following manner. The idea is to replace the original goal with a state visited by the failed episode. As the reward function remains unchanged, this change of goals hints the agent how to achieve the new goal in the environment. HER assumes that the mechanism of reaching the new goal helps the agent learn for the original goal.
 In our experiment, the reward is $0$ if the goal is achieved, $-1$ otherwise.

 \subsection{Rollback Mechanism} \label{sec:rollback}
 As shown in Section \ref{sec:exp_stage1} and \ref{sec:exp_stage2}, neither cube rotation nor layer-wise operation can achieve a success rate of $100\%$ on their sub-tasks. The cube operators, including cube rotation and layer-wise operation, are performed sequentially, constituting an open loop system without any feedback signals. This results in the success rate of the whole task exponentially decreasing with the number of actions.
 One situation leading to the failure is that Stage 1 fails to re-pose the cube correctly, causing a wrong layer is rotated in Stage 2. Another situation is Stage 2 does not rotate the target layer to the target angle, which will hinder performing the following operations.
 %the cube solver is not able to plan a path since there is no operation with rotation angle less than 90$^o$. 
 
\begin{figure}
     \centering
     \includegraphics[width=\linewidth]{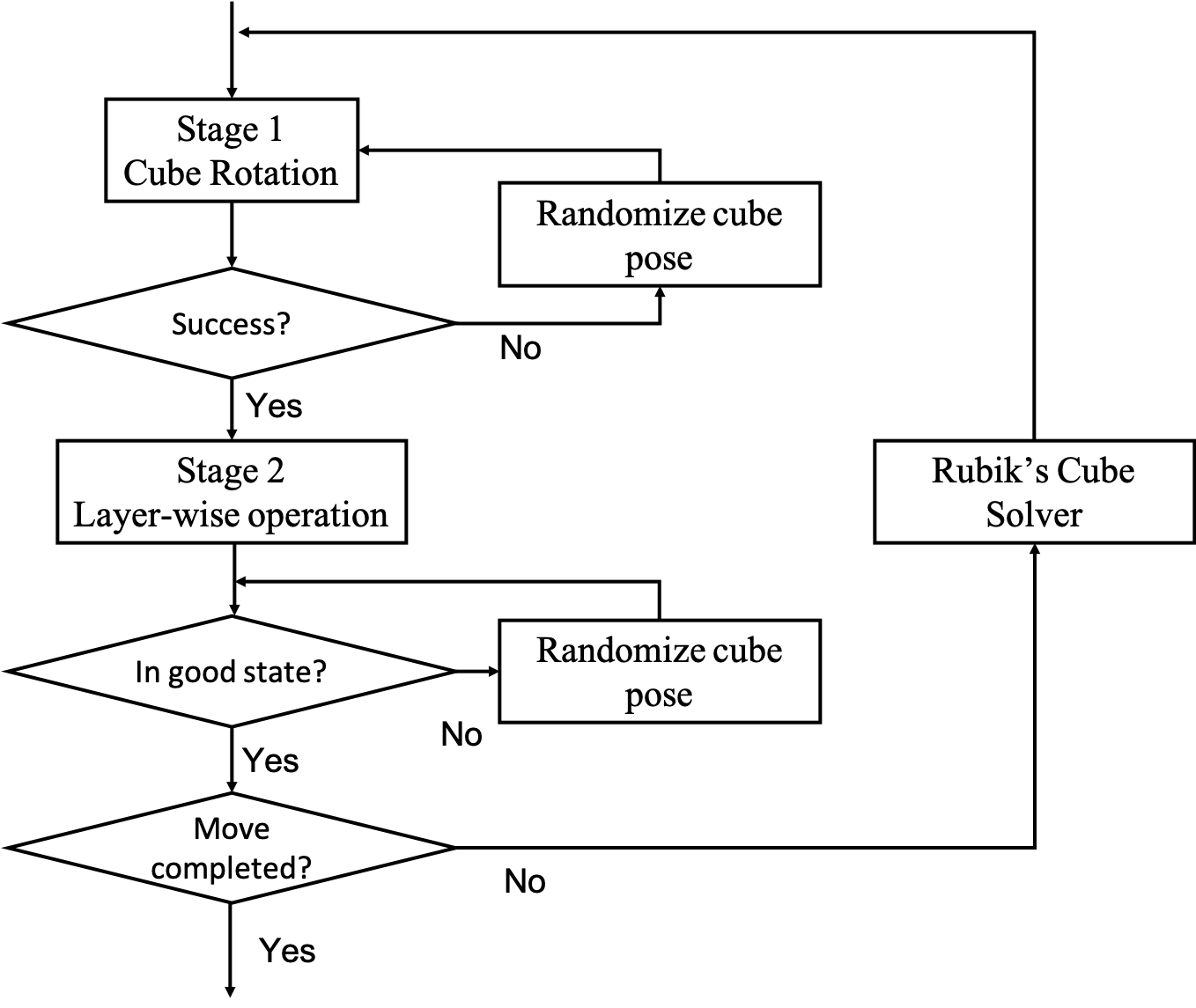}
     \caption{Work flow of the rollback mechanism. First check the cube status after Stage 1 and re-do Stage 1 if it fails. Then check whether the cube is in a good state after Stage 2 and perform randomization if it is not in a good state. Finally check whether the move is completed to determine whether to re-plan a new move sequence.}
     \label{fig:rollback}
 \end{figure}

 In light of this, we develop a rollback mechanism to improve system stability.
 The workflow of the rollback mechanism is shown in Fig. \ref{fig:rollback}. After Stage 1, the pose of the Rubik's Cube will be checked. If it deviates from the goal pose, we will first rotate it to a random pose and then repeat Stage 1. Applying randomization can help avoid the cube falling into the local minimum. Similar to Stage 1, we check whether the Rubik's Cube is in a good state after performing the layer-wise operation, i.e. all cubelets align with each other. If not, we first restore it into a good state by applying randomization. Although a separate model can be trained to restore the cube's state, we experimentally find it performs well by applying randomization. The final step is checking whether the desired move is completed to determine whether to re-plan a new sequence.   

\section{EXPERIMENTS} \label{sec:experiment}
The performance of the proposed method is verified in Mujoco simulator \cite{todorov2012mujoco}. First the success rate of two atomic actions is evaluated in Section \ref{sec:exp_stage1} and \ref{sec:exp_stage2}. 
To test our method, we generate $1400$ randomly scrambled Rubik's Cubes. The results show that our structure can achieve $90.3\%$ success rate in this task. To ensure reproducibility and further innovations, we will release our code and models upon publication.

 \subsection{Experiment Setup}
 %To solve a 2x2x2 Rubik's Cube using a Shadow Hand, we implement the whole environment in Mujoco simulator as described below.
 We implement 2x2x2 Rubik's Cube environments in Mujoco simulator as described below.
 
 \subsubsection{Shadow Hand}
 We use a highly dexterous human-sized manipulator, Shadow Hand, in our task. Shadow Hand is a 24-DoF anthropomorphic manipulator, where the first, middle and ring fingers have 3 actuated joints and 1 under-actuated joint and little finger and thumb have 5 actuated joints. Besides, the wrist has 2 actuated joints. In our experiments, the 20 actuated joints are controlled based on position control.
 
 \subsubsection{Simulator}
 Our experiments are conducted in the Mujoco physics simulator \cite{todorov2012mujoco}, which has a stable contact dynamics \cite{erez2015simulation} and is suitable for rich-contact hand manipulation tasks.
 
 \subsubsection{Rubik's Cube}
 We implement a 2x2x2 Rubik's Cube in the simulator. Eight cubelets are of 2.5x2.5x2.5cm$^3$ and connected by a ball joint located at the cube center. Besides, there are circular chamfers on the edges of each cubelet so that the Rubik's Cube has an angle tolerance around $5^\circ$. For example, if there is an angle within $5^\circ$ between up layer and down layer, we are still able to perform $R$ or $F$.
 
    \begin{table}[]
     \centering
     \caption{Observations of cube rotation environment (denoted as CubeRotEnv) and layer-wise operation environment (denoted as LayerOpEnv), respectively.}
     \resizebox{\linewidth}{!}{
        \begin{tabular}{llcc}
            \hline \hline 
             Name & Dim. & CubeRotEnv & LayerOpEnv  \\
            \hline
            Hand joints angles & 24D & \checkmark & \checkmark \\
            Hand joints velocities & 24D & \checkmark & \checkmark \\
            Cube position & 3D & \checkmark & \checkmark \\
            Cube orientation & 4D  & \checkmark & \checkmark \\
            Cube velocity & 3D & \checkmark & \checkmark \\
            Cube angular velocity & 3D & \checkmark & \checkmark \\
            Layer angle difference & 1D & \xmark & \checkmark \\
            \hline
        \end{tabular} }
     \label{tab:specification}
 \end{table}
 
 \subsection{Environments}
 We develop two environments for the problem. The two environments correspond to two stages, respectively. The actions are 20-dim joint angles and the observations are listed in TABLE \ref{tab:specification}.
 
 \subsubsection{Cube rotation environment}
\begin{figure}
     \centering
     \subfloat[]{\includegraphics[width=120pt]{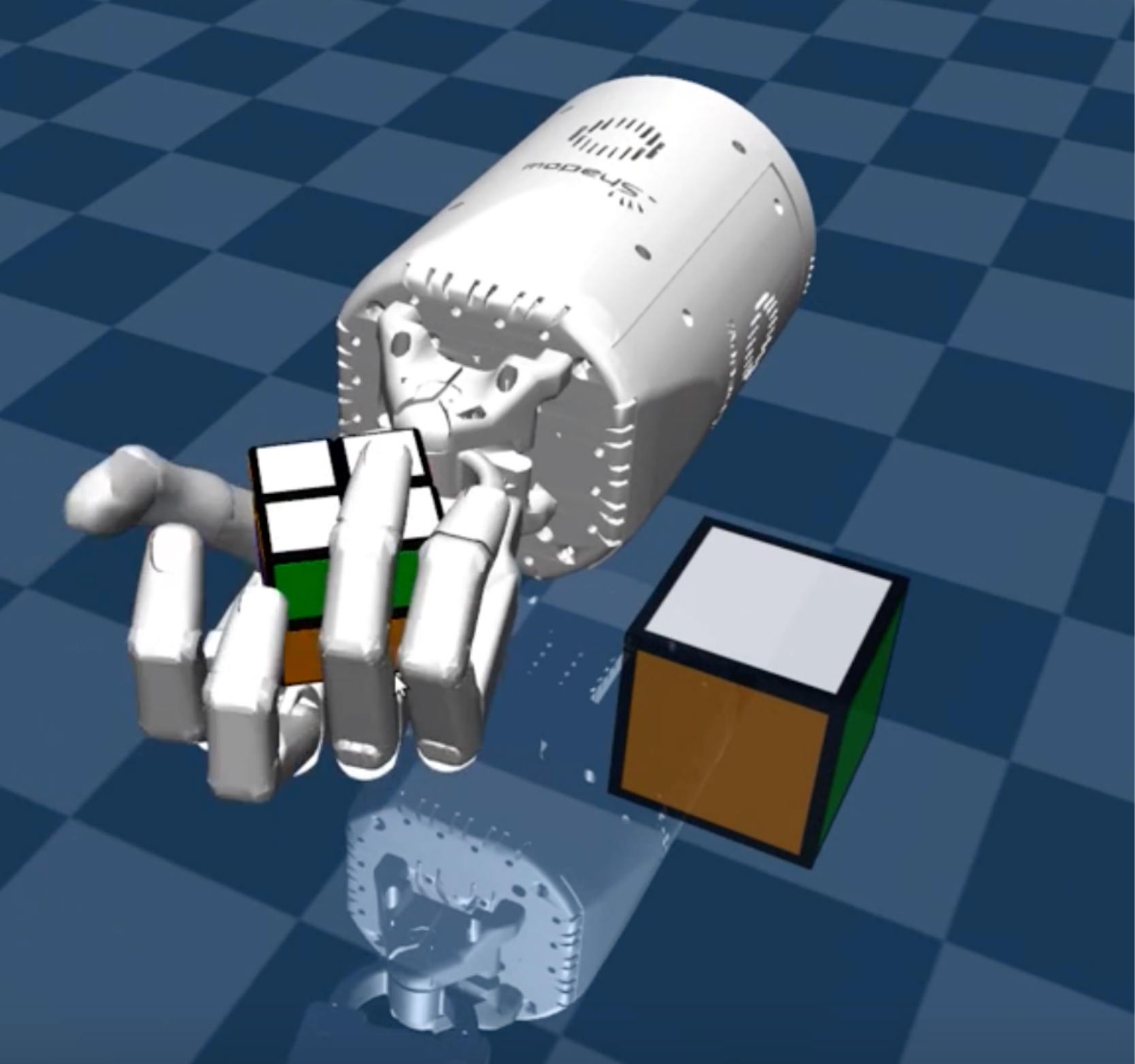}}\ 
     \subfloat[]{\includegraphics[width=120pt]{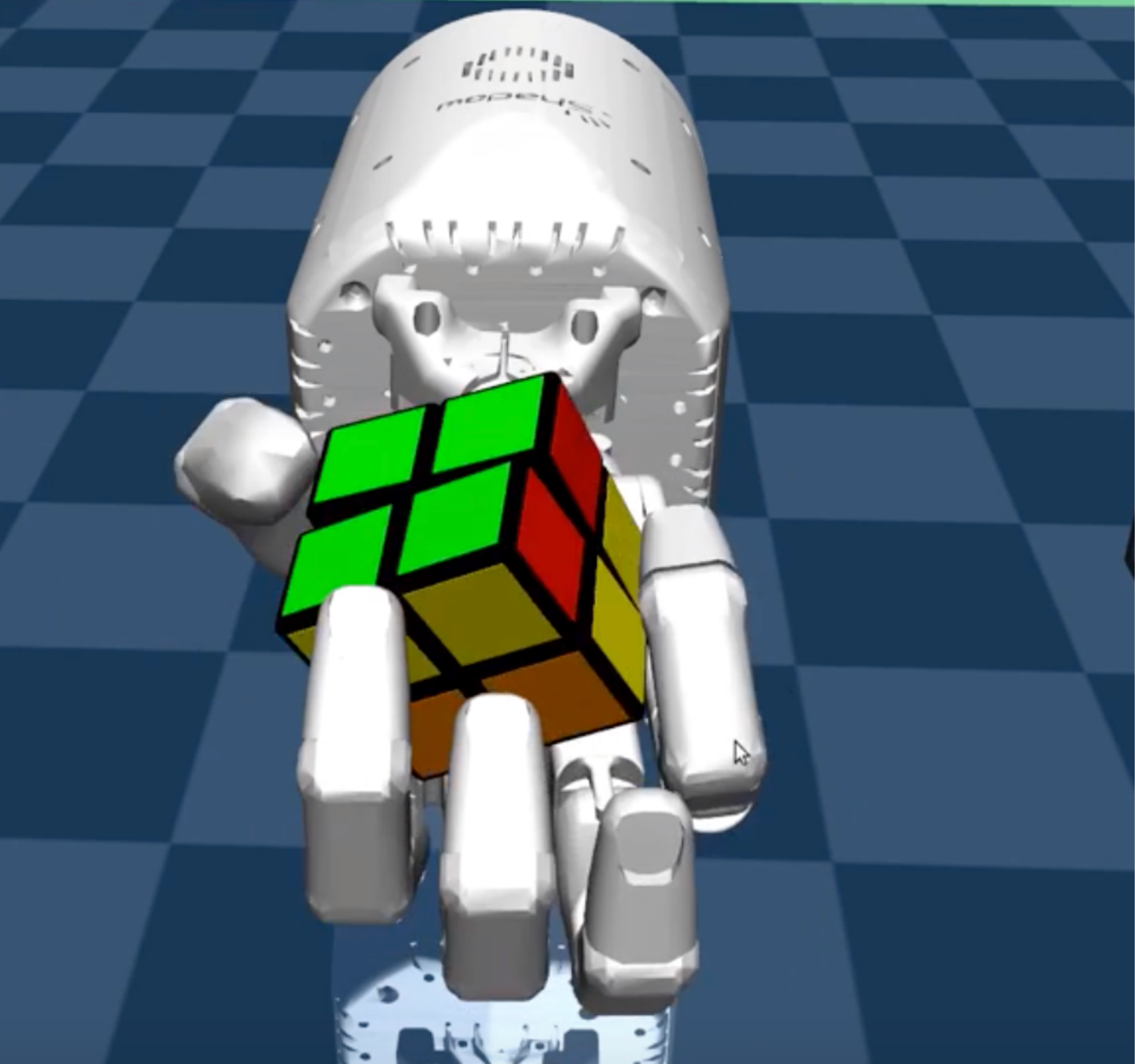}}
     \caption{Illustration of our environments. (a) CubeRotEnv. The Rubik's cube is placed on the palm and the target orientation is indicated on the right. The goal is a 7-dim vector including the target position and target rotation. (b) LayerOpEnv. The goal is the target angle between two layers.}
     \label{fig:env-1}
 \end{figure}

 In the cube rotation task (CubeRotEnv), a block is placed on the palm of the hand. The objective is to  manipulate the block to achieve a target pose. The goal is a $7$-dim vector including the target position (in Cartesian coordinates) and target rotation (in quaternions) and the achieved goal is the pose of the block. The environment is shown in Fig.~\ref{fig:env-1} (a).
 
 \subsubsection{Layer-wise operation environment}
 In the layer-wise operation task (LayerOpEnv), a block is placed on the palm of the hand. The block contains two layers. The objective is to  manipulate the two layers to achieve a target angle. The goal is the angle between the target layers and the achieved goal is the angle between the two layers. The environment is shown in Fig.~\ref{fig:env-1} (b).
% \begin{figure}
%      \centering
%      \includegraphics[width=140pt]{./fig/rubik-env-1.pdf}
%      \caption{LayerOpEnv.}
%      \label{fig:env-2}
%  \end{figure}
  
 \subsection{Cube Rotation} \label{sec:exp_stage1}
\begin{figure}
     \centering
     \includegraphics[width=\linewidth]{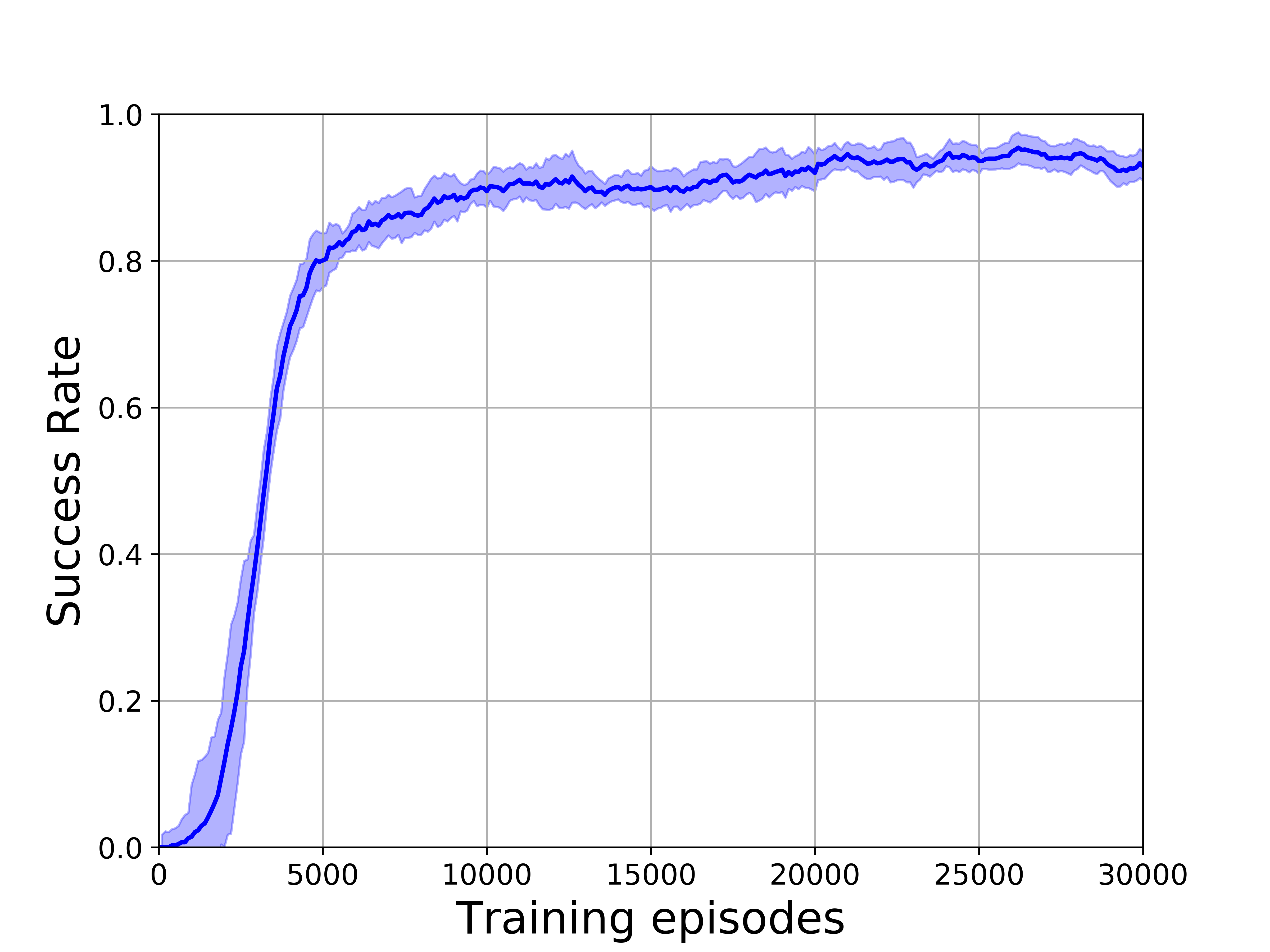}
     \caption{Success rate of cube rotation. The shaded area represents one standard deviation. Our model achieves a success rate of over $90\%$ at the training stage.}
     \label{fig:stage1}
 \end{figure}

 We show the performance of the atomic actions in this part. 
 We employ Deep Deterministic Policy Gradient (DDPG) \cite{lillicrap2015continuous} in our cube rotation environment. HER is also employed where the episodes failed reaching the goal are stored in replay buffer with a modified goal, aiming to learn from failed experience. The network contains 3 hidden layers each with $256$ units and each episode lasts 100 steps. At the beginning of each episode, the cube is randomly initialized on the hand palm with an arbitrary orientation. The position threshold $\delta_x$ is $0.01$ m and the orientation threshold $\delta_q$ is $0.1$ rad. The success rate at the training stage is shown in Fig. \ref{fig:stage1}. It shows our model can achieve a stable success rate of over $90\%$. After training for $30,000$ episodes, we test our model for $1000$ times and it achieves an average success rate of $95.2\%$.
 
  \begin{table*}[]
    \centering
    \caption{The success rate (SR) and action number (AN) of our structure and the non-rollback version conditioned on move distance. The results are tested over $100$ scrambled Rubik's Cubes for each move distance. Our structure achieves an average success rate over all move distance of $90.3\%$.}
    \resizebox{\textwidth}{!}{
    \begin{tabular}{l|cccccccccccccc }
        \hline
        Move Distance & 1 & 2 & 3 & 4 & 5 & 6 & 7 & 8 & 9 & 10 & 11 & 12 & 13 & 14\\
        \hline
        SR (Rollback) & 1.0 & 1.0 & 0.95 & 0.95 & 0.89 & 0.95 & 0.94 & 0.86 & 0.85 & 0.89 & 0.8 & 0.8 & 0.88 & 0.88 \\
        SR (Non-rollback) & 0.93 & 0.83 & 0.82 & 0.74 & 0.72 & 0.62 & 0.46 & 0.43 & 0.42 & 0.43 & 0.34 & 0.41 & 0.37 & 0.46 \\
        \hline
        AN (rollback) & 4.8 & 9.11 & 11.49 & 14.80 & 18.04 & 22.22 & 26.87 & 28.39 & 34.91 & 36.94 & 41 & 41 & 41.81 & 39.65\\
        AN (Non-rollback) & 4.81 & 6.07 & 9.90 & 12.81 & 15.22 & 19.55 & 21.57 & 25.11 & 26.06 & 29.95 & 34.2 & 34.15 & 34.65 & 32.60 \\
        \hline
    \end{tabular}}
    \label{tab:sr_an}
\end{table*}
 
 \subsection{Layer-wise Operation} \label{sec:exp_stage2}
 As discussed in Section \ref{sec:atomic_actions}, the difficulty level of the six generalized moves $\{U,U',R,R',F,F'\}$ varies with respect to an anthropomorphic hand. In this part we will quantitatively compare them.
 
\begin{figure} 
     \centering
     \includegraphics[width=\linewidth]{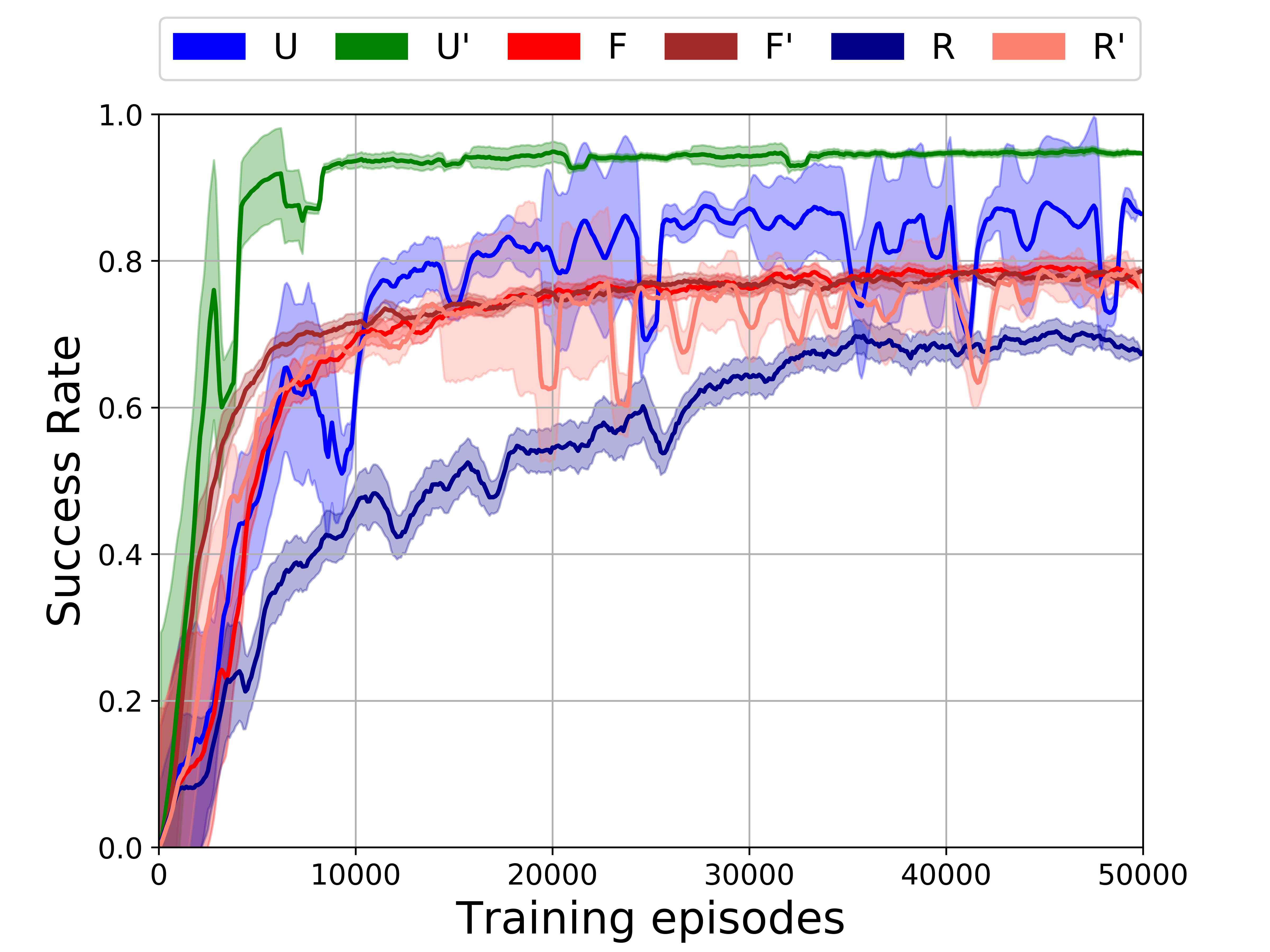}
     \caption{The success rate of 6 moves. The shaded area represents a standard deviation. Among all 6 moves, $U'$ (the green line) performs the best with the highest and the most stable success rate. }
     \label{fig:stage2}
   \end{figure}

 Similar to Section \ref{sec:exp_stage1}, DDPG and HER are employed to train six models corresponding to the six moves. The network contains 3 hidden layers each with $256$ units. The angle threshold $\delta_\theta$ is $0.1$ rad. Fig.~\ref{fig:stage2} shows the success rate at the training stage. It is evident that the move $U'$ outperforms all other moves. It has a faster training speed compared with $R$ and $U$, and more stable performance, compared with $R'$ and $U$. Although $F$ and $F'$ also have stable success rates, it is much lower than $U'$. Therefore, the move $U'$ is selected as the action of Stage 2. We test the move $U'$ for $1000$ times with randomized block poses and it achieves an average success rate of $92.3\%$.
 
%  \subsection{Joint Training}
%   \begin{itemize}
%       \item The whole process is MDP, where the state is the sub-task.
%       \item One advantage of our approach is that the process of bad behavior is very fast.
%       \item The hand is not systematic, thus the difficulty among different rotated layers, the rotation directions will vary significantly.
%   \end{itemize}

 \subsection{Solving the Rubik's Cube} \label{sec:exp_success}
 To comprehensively evaluate the performance of the proposed model, we test the success rate conditioned on the minimum moves needed to restore a Rubik's Cube, which we refer to as move distance. The move distance reflects the difficulty level of solving a scrambled cube. For a 2x2x2 Rubik's Cube, the move distance ranges from $1$ to $14$. So we first collect $1400$ scrambled samples with $100$ in each move distance and test our model to solve these cubes. We compare our method with a non-rollback version to illustrate the effectiveness of the rollback mechanism. The result is shown in TABLE \ref{tab:sr_an}. It shows that the Success Rate (SR) of both structures decreases with move distance. The rollback mechanism has a significant effect on improving SR since the SR(rollback) is higher than non-rollback version under every move distance. Our structure achieves an average success rate of $90.3\%$ with the lowest SR of $80\%$ at the move distance of $11$ and $12$. 
 
 Since the rollback mechanism involves additional actions to restore from abnormal status, we record the average Action Number (AN) as shown in TABLE \ref{tab:sr_an}. AN refers to the number of atomic actions it involved in restoring a cube.  AN (rollback) is larger than AN (non-rollback) under each move distance. But considering the significant improvement on SR, such cost is still acceptable.

%   \begin{figure}
%      \centering
%      \includegraphics[width=220pt]{./fig/success_whole.png}
%      \caption{Pick a better one.}
%      \label{fig:success_rate}
%  \end{figure}
%  \subsection{Ablation analysis}
%   \begin{itemize}
%       \item effect of $-pi/2*1.2, pi/2*1.2$
%       \item distribution of end state (pce visualization), figure show hand locations.
%   \end{itemize}
% \section{CONCLUSIONS}

\section{Conclusions}
In this paper, we have presented a hierarchical learning method to solving a 2x2x2 Rubik's cube using a Shadow hand. Our method employs a model-based Rubik's Cube solver to plan a move sequence, and model-free atomic actions to control the hand to accomplish the planned moves. The capability of our model is demonstrated in Mujoco simulation platform and it achieves a $90.3\%$ success rate in solving $1400$ scrambled cubes. 

We see several future directions to extend our work. One  is  to jointly optimize the two atomic actions. Currently, they are separately trained which may result in problematic end states. Problematic end states refer to the end states of one stage are too difficult for the next stage, e.g., the hand holds the cube with its thumb and first finger. This can be attributed to that each stage only considers its own goal. One possible solution is to define a goal with a longer horizon, i.e. the goal of Stage 1 also considers whether its end states are too challenging for Stage 2 to start with.  We may also train a separate classifier to infer the difficulty level the the state.

Another direction is to deploy our method on real hardware systems. To mitigate the gap between the simulator and the reality, we may consider sim2real techniques including extensive domain randomization and memory augmented control polices. Besides, human demonstrations can also be employed to reduce sample complexity while obtaining natural and robust movements.
 
\section{Acknowledgment}
Research presented in this paper is partially supported by the Hong Kong RGC grant No. 14200618 awarded to Professor Max Q.-H. Meng.

\bibliographystyle{IEEEtran}
\bibliography{iros2019}
\end{document}